\documentclass[runningheads]{llncs}
\usepackage{graphicx}
\usepackage{multirow}
\begin{document}
\title{Midpoint Regularization: from   High Uncertainty Training   to     Conservative Classification }
\titlerunning{Midpoint Regularization}
%
\author{Hongyu Guo 
}
\authorrunning{Hongyu Guo}
%
\institute{National Research Council Canada \\ 
1200 Montreal Road, Ottawa, Ontario, K1A 0R6  \\ 
\email{hongyu.guo@nrc-cnrc.gc.ca}\\
}
%
\maketitle              
\begin{abstract}
Label Smoothing (LS) improves model  generalization  through penalizing  models from generating overconfident output distributions. For each training sample the LS strategy smooths the one-hot encoded training signal by distributing its distribution mass over the non-ground truth classes. We extend this technique by considering example pairs, coined  PLS. PLS first creates midpoint samples by averaging random sample pairs and then  learns a smoothing distribution during training for each of these midpoint samples, resulting in midpoints with high uncertainty labels for training. We empirically show that PLS significantly outperforms LS, achieving up to 30\% of relative classification error reduction. We also visualize  that PLS  produces very low winning softmax scores for both in and out of distribution samples. 
\end{abstract}

\section{Introduction}
Label Smoothing (LS) is a commonly used output distribution regularization technique to improve the generalization performance of deep learning models~\cite{Chorowski2016TowardsBD,Huang2019GPipeET,structurels,Real2019RegularizedEF,Szegedy2016RethinkingTI,Vaswani2017AttentionIA,8579005}. 
Instead of training with data associated with one-hot labels, models with label 
smoothing  are trained on samples with  soft targets, where each target is  
a weighted mixture of the ground truth one-hot label  with the uniform distribution of the classes. 
This penalizes overconfident output distributions, resulting in improved model generalization~\cite{Lukasik2020DoesLS,MullerKH19,DBLP:conf/iclr/PereyraTCKH17,yuan2019revisit,DBLP:conf/aaai/ZhuJZGHSZ20}. 

\begin{figure}[h]
\caption{
Average  distribution mass (Y-axis) of PLS and Mixup (from sample pairs with  different labels) and ULS on Cifar10's 10 classes (X-axis). 
 PLS' class-wise training signals  are  spread out,  with  gold labels probabilities  close to 0.35, which has  larger  uncertainty than that of 0.5 and 0.9 from Mixup and ULS, receptivity. 
 }
\label{fig:trainsignal}
\centering
\includegraphics[
width=4.25972in
]{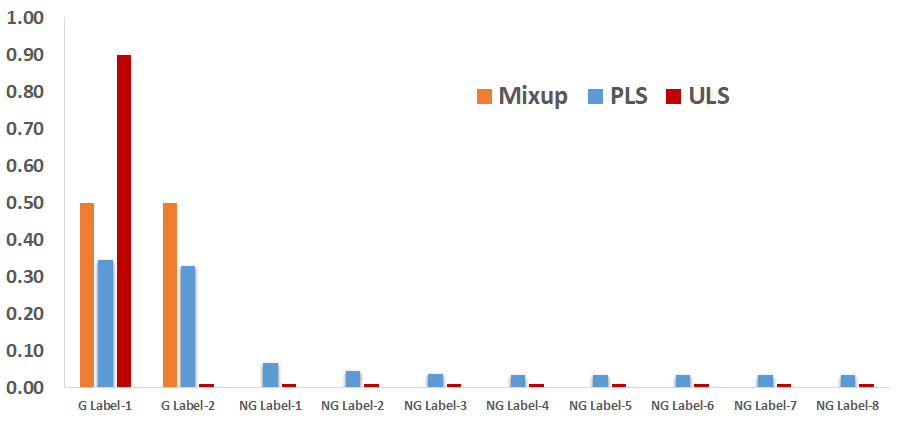}
\end{figure}

\begin{figure}[h]
\caption{
Histograms of the winning softmax scores on Cifar10 validation data, generated by  ULS  (right Y-axis) and by Mixup and PLS (left  Y-axis) where the X-axis depicts the softmax score. The PLS produces very low softmax winning scores; the differences versus the ULS and Mixup  are striking: the PLS  models are extremely sparse  in the high confidence region at the right end.}
\label{fig:pilot}
\centering
\includegraphics[
width=3.972in
]{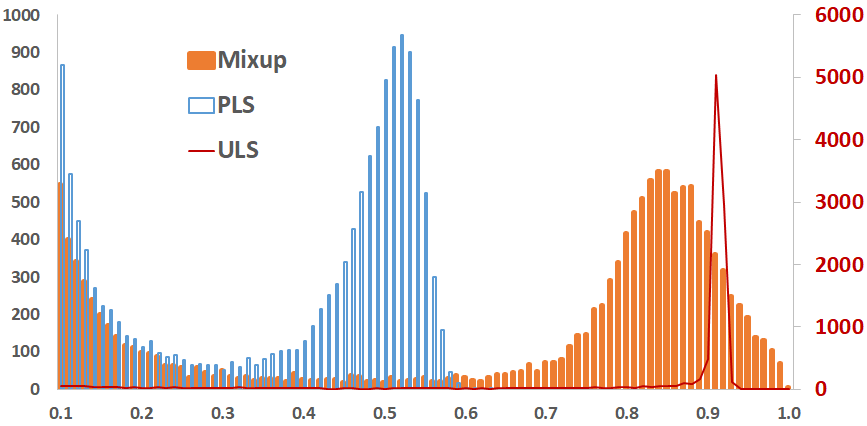}
\end{figure}

When smoothing the one-hot training signal, existing LS methods, however, only consider the distance between the only gold label and the  non-ground truth targets.  
This motivates our Pairwise Label Smoothing (PLS) strategy, which takes  a pair of samples as input. 
In a nutshell, the PLS first creates a midpoint sample by averaging the inputs and labels of a sample pair,  
and then distributes the  distribution mass of the two ground truth  targets  of the new midpoint sample over its non-ground truth classes. 
Smoothing with a pair  of ground truth labels  enables  PLS to  preserve  the relative distance between the two truth labels  while being able to further soften that between the truth labels  and the other class targets. 
Also, unlike current LS methods, which typically require to  find a global smoothing distribution mass through cross-validation search, PLS  automatically  learns the   distribution mass for each input pair during training.
Hence, it effectively eliminates the   turning efforts for searching the right level of smoothing strength 
for different applications.

Also, smoothing with a pair of labels empowers PLS   to be trained with low  distribution mass (i.e., high uncertainty)  for the ground truth  targets. 
Figure~\ref{fig:trainsignal} depicts the average  values  (over all the midpoint  samples created from sample  pairs with different labels) of the 10 target classes in Cifar10 used by PLS for training. These ground truth targets are smaller than 0.35, which represents  much larger  uncertain than the 0.5 and 0.9 in Mixup and the  uniform label smoothing (denoted as ULS), respectively. 
Owing to such   high uncertainty ground truth  target training signals, 
PLS  produces extremely low winning softmax scores during testing, resulting in very conservative classification decisions.  Figure~\ref{fig:pilot} depicts the histograms of the winning  softmax scores of Mixup, ULS 
and PLS on the Cifar10 validation data. 
Unlike Mixup and ULS,  PLS  produced  very conservative  softmax scores,  with extremely sparse distribution in  the high confidence region. 
Such conservative classifications  are produced not only  on in-distribution data but also on out-of-distribution samples. 

We empirically show that 
 PLS  significantly 
outperforms  LS and  Mixup~\cite{Mixup17}, 
with up to 30\% of relative classification error reduction. 
We  also visualize that PLS produces extremely conservative predictions  in testing time, 
with many of its  winning predicted softmax scores  slightly over 0.5 for both in and out of distribution samples.

\section{Related Work}
Label smoothing has shown to  provide consistent model generalization gains across many tasks~\cite{Szegedy2016RethinkingTI,DBLP:conf/iclr/PereyraTCKH17,7780883,MullerKH19,Lukasik2020DoesLS,structurels}. 
Unlike the above methods which apply  label smoothing to  each single input, our smoothing strategy leverages a pair of inputs for label smoothing. Furthermore, 
unlike existing LS approaches which deploy one static and uniform smoothing distribution for all the training samples, our strategy  automatically learns a dynamic distribution mass for each sample during training. 

Mixup methods~\cite{DBLP:journals/corr/abs-1906-06875,DBLP:conf/aaai/Guo20,DBLP:journals/corr/abs-1905-08941,adamixup,DBLP:conf/iclr/TokozumeUH18,cutmix19,Mixup17
} 
 create  {\em a large number} of synthetic samples with various features from a  sample pair,  through  interpolating   the pair's   both features and labels  with mixing coefficients randomly sampling  between [0,1]. In contrast, 
our approach   creates {\em only the midpoint} sample of a sample pair (equivalent to  Mixup with a fixed  mixing ratio of 0.5).
More importantly, our method adds a label smoothing component on the midpoint samples. That is, we 
adaptively learn  a  distribution mass to smooth the pair of ground truth  labels of each midpoint sample, 
 with the aims of  deploying   high uncertainty training targets (much higher than that of Mixup) for output distribution regularization. Such  high uncertain training labels  result in PLS producing very conservative classification decisions and superior  accuracy than  Mixup.

The generation of the  smoothing distribution  in our method is also related to  self-distillation~\cite{ahn2019variational,Furlanello2018BornAN,hinton2015distilling,mobahi2020selfdistillation,DBLP:conf/aaai/YangXQY19}. However, these  approaches treat the final predictions as target labels for a new round of training, and  the teacher and student architectures are identical~\cite{mobahi2020selfdistillation}.  In our method, the classification   and  the smoothing distribution  generator   have different network architectures, and the training targets for the classification  are a mixture of the outputs of the two networks.

\section{Label Smoothing over Midpoint Samples}
Our proposed  method PLS leverages a pair of samples,  randomly selected from the training data set, to conduct label smoothing. It first creates a midpoint sample for each sample pair, and then adaptively learns a smoothing distribution for each midpoint sample. These midpoint samples are then used for training to regularize the learning of the networks. 
\subsection{Preliminaries}
We consider a standard classification setting with a given training data set $(X; Y)$ associated with $K$ candidate classes $\{1,2,\cdots,K\}$. 
For an example $x_i$ from the training dataset $(X; Y)$, we denote the ground truth distribution $q$ over the labels as $q(y|x_i)$ ( $\sum_{y=1}^{K} q(y|x_i) = 1$). Also, we denote  a neural network  model to be trained as $f_{\theta}$ (parameterized with $\theta$), and it produces 
a conditional  label distribution over the $K$ classes as $p_{\theta}(y|x_i)$:
\begin{equation}
p_{\theta}(y|x_i) = \frac{\exp(z_{y})}{\sum_{k=1}^{K}\exp(z_{y_{k}})}, 
\end{equation}
with 
 $\sum_{y=1}^{K} p_{\theta}(y|x_i) = 1$, and $z$
is noted as the logit of the model $f_{\theta}$. The logits are generated with two steps: 
 the model $f_{\theta}$ first  constructs the $m$-dimensional input embedding $S_{i} \in R^{m}$ for the given input $x_{i}$, and then passes it 
  through
a linear   fully connected layer $W_{l} \in R^{K \times m}$: 
\begin{equation}
\label{logit}
S_{i} = f_{\theta}(x_{i}), 
z=  W_{l}S_{i}.
\end{equation}
During learning, the model $f_{\theta}$ is trained to optimize the parameter $\theta$ using the $n$ examples from  $(X; Y)$ by  minimizing the cross-entropy loss: 
\begin{equation}
\ell =- \sum_{i=1}^{n} H_i(q,p_{\theta}). 
\end{equation} 
Instead of using one-hot encoded vector for each example $x_{i}$ in $(X; Y)$, label smoothing (LS) adds a smoothed label distribution  (i.e., the prior distribution) $u(y|x_{i})$ to each example $x_{i}$, forming a new target label, namely  soft label:  
\begin{equation}
\label{oriequ}
q^{\prime}(y|x_i) = (1-\alpha) q(y|x_i) + \alpha u(y|x_i), 
\end{equation}
where hyper-parameter $\alpha$ is a weight factor ($\alpha \in [0,1]$) needed to be tuned to indicate the smoothing strength for the one-hot label. This modification results in a new loss:
\begin{equation}
\label{lsloss}
\ell^{\prime} 
= - \sum_{i=1}^{n} \Big[ (1-\alpha) H_i(q,p_{\theta}) + \alpha H_i(u,p_{\theta}) \Big]. 
\end{equation} 
Usually, the $u(y|x_{i})$ is a uniform distribution, independent of data $x_{i}$, as $u(y|x_{i}) = 1/K$, and hyper-parameter $\alpha$ is tuned with cross-validation.

\subsection{Midpoint Generation}
\label{plsmethod}
For a sample, denoted as $x_{i}$, from the provided training  set $(X; Y)$ for training, PLS first randomly selects~\footnote{For efficiency purpose, we implement this by randomly selecting a sample from the same mini-batch during training.} another training sample $x_{j}$. For the  pair of samples $(x_{i}; y_{i})$ and $(x_{j} ; y_{j})$, where $x$ is the
input and $y$ the one-hot encoding of the corresponding class, PLS then generates a synthetic sample  through element-wisely averaging both the input features and the labels, respectively, as follows:  
\begin{equation}
x_{ij} =  (x_{i} +  x_{j})/2, 
\end{equation}
\begin{equation}
\label{equ:2}
q(y|x_{ij}) = ( y_{i} +  y_{j})/2.  
\end{equation}
These synthetic samples can be considered as the  {\em midpoint samples} of the original sample pairs. It is also worth noting that, the midpoint samples here are equivalent to fixing the linear interpolation mixing ratios as 0.5 in the Mixup~\cite{Mixup17} method. 
Doing so, for the midpoint sample $x_{ij}$ we have the ground truth distribution $q$ over the labels as $q(y|x_{ij})$ ($\sum_{y=1}^{K} q(y|x_{ij}) = 1$). 
The newly resulting  midpoint  $x_{ij}$ will  then be  used for label smoothing (will be discussed in detail in Section~\ref{evovinglabel}) before feeding into the networks for training. 
In other words, the predicted logits as defined in Equation~\ref{logit} is computed by first generating the  $m$-dimensional input  embedding $S_{ij} \in R^{m}$ for  $x_{ij}$ and then passing through the fully connected linear layer to construct the logit $z$: 
\begin{equation}
S_{ij} = f_{\theta}(x_{ij}), 
\end{equation}
\begin{equation}
\label{mmix22}
z= W_{l}S_{ij}.
\end{equation}
Hence, the predicted conditional  label distribution over the $K$ classes 
produced by the networks is as follows:
\begin{equation}
p_{\theta}(y|x_{ij}) = \frac{\exp(z_{y})}{\sum_{k=1}^{K}\exp(z_{y_{k}})}. \end{equation}

\subsection 
{Learning Smoothing Distribution for Midpoints}\label{evovinglabel} 
PLS leverages a learned distribution,  which depends on the input $x$, to dynamically generate the smoothing distribution mass for midpoint samples to distribute their ground truth target distribution to the non-target classes. 
To this end, the PLS implements this by adding a  fully connected layer  to the network $f_{\theta}$. That is,   the 
$f_{\theta}$ produces two projections from the {\em penultimate} layer representations of the network:  one for the logits as the original network (Equation~\ref{mmix22}), and another for generating the smoothing distribution as follows. 

In specific, an additional  fully connected
layer $W_{t} \in R^{K \times m}$ is added to the original networks $f_{\theta}$ to produce the smoothing distribution over the $K$ classification classes. 
That is, for the given 
input image $x_{ij}$,  its smoothing  distributions over the $K$ classification  targets, denoted as  $u^{\prime}_{\theta}(y|x_{ij})$, are computed as follows:  
\begin{equation}
\label{prelabel}
u^{\prime}_{\theta}(y|x_{ij}) = 
\frac{\exp(v_{y})}{\sum_{k=1}^{K}\exp(v_{y_{k}})},  
\end{equation}
   \begin{equation}
   \label{sha}
   v = \sigma (W_{t}S_{ij}), 
   \end{equation}
where $\sigma$ denotes the Sigmoid function, and  
 $S_{ij}$ is the same input  embedding as that in Equation~\ref{mmix22}. In other words, the two predictions (i.e., Equations~\ref{mmix22} and~\ref{sha}) share the same networks except the last fully connected layer. 
That is, the only difference between PLS and the original networks is the added  fully connected layer $W_{t}$. 
The added Sigmoid function here aims to  squash the smoothing distributions learned for different targets to the same range of  [0, 1]. 

After having the smoothing distributions $u^{\prime}_{\theta}(y|x_{ij})$, PLS then uses them to smooth the ground truth labels  $q(y|x_{ij})$ as described in Equation~\ref{equ:2}, with   average:  
\begin{equation}
\label{merge}
q^{\prime}(y|x_{ij}) = ( q(y|x_{ij}) +  u^{\prime}_{\theta}(y|x_{ij}))/2. 
\end{equation}

The loss function of PLS thus becomes the follows:
\begin{equation}
\ell^{\prime} 
= - \sum_{i=1}^{n} \Big[ 0.5 \cdot  H_i(q(y|x_{ij}),p_{\theta}(y|x_{ij})) 
+  0.5\cdot H_i(u^{\prime}_{\theta}(y|x_{ij}),p_{\theta}(y|x_{ij})) 
\Big].
\end{equation} 

Coefficient 0.5 here  helps prevent the network from over-distributing its label distribution  to the non-ground truth targets. Over-smoothing  degrades the model performance, as will be shown in the experiments.
Also, 
0.5 here   causes the resulting training signals to have high  uncertainty regarding the ground truth targets. 
Such high uncertainty  helps train  the models to make conservative predictions. 

\subsection{Optimization}
For training, PLS minimizes, with gradient descent on mini-batch, the  loss  $\ell^{\prime}$. 
One more issue needs to be addressed for the training. That is, the midpoint samples used for training, namely $(x_{ij}; y)$, may lack  information on the original training samples  $(x_{i}; y)$ due to the average operation in midpoint generation. To compensate this, 
we alternatively feed inputs to the networks with either a mini-batch from the original inputs, i.e., ${x}_{i}$ or ${x}_{ii}$,  or a mini-batch from the midpoint samples, i.e., ${x}_{ij}$. 
Note that, when training with the former, the networks still  need to learn to assign the smoothing distribution $ u^{\prime}_{\theta}(y|x_{ii})$ to form the  soft  targets $q^{\prime}(y|x_{ii})$ for the sample ${x}_{ii}$. 
As will be shown in the experiment  section, this training strategy is   important to PLS'  regularization effect. 

\begin{table*}[h]
  \centering
\scalebox{0.9}{
\begin{tabular}{l|c|c|c|c|c|c}\hline
Methods& MNIST&Fashion&SVHN&Cifar10&Cifar100&Tiny-ImageNet\\ \hline
PreAct ResNet-18 &0.62	$\pm$0.05&4.78$\pm$0.19&3.64$\pm$	0.42&5.19$\pm$0.30&24.19$\pm$	1.27&39.71$\pm$0.08\\
ULS-0.1 &0.63$\pm$0.02&4.81$\pm$0.07&3.20 $\pm$0.06&4.95$\pm$0.15&21.62$\pm$0.29&38.85$\pm$0.56\\
ULS-0.2&0.62$\pm$0.02&4.57$\pm$0.05&3.14$\pm$0.11&4.89$\pm$0.11&21.51$\pm$0.25&38.54$\pm$0.32\\
ULS-0.3&0.60$\pm$0.01&4.60$\pm$0.06&3.12$\pm$0.03&5.02$\pm$0.12&21.64$\pm$0.27&38.32$\pm$0.37\\
\hline
Mixup  &0.56$\pm$0.01&4.18$\pm$	0.02&3.37$\pm$0.49&3.88$\pm$	0.32&21.10$\pm$0.21&38.06$\pm$0.29\\
Mixup-ULS0.1 &0.53$\pm$0.02&4.13$\pm$0.10&2.96$\pm$0.39&4.00$\pm$0.17&21.51$\pm$0.51&37.23$\pm$0.48\\
Mixup-ULS0.2 &0.53$\pm$0.03	&4.18$\pm$0.09	&3.02$\pm$0.42&3.95$\pm$0.13&21.41$\pm$0.55&38.21$\pm$0.38\\
Mixup-ULS0.3  &0.50$\pm$0.02	&4.15$\pm$0.06	&2.88$\pm$0.31&4.06$\pm$0.04&$20.94\pm$0.49&38.93$\pm$0.43\\
\hline
PLS &\textbf{0.47 $\pm$0.03}&\textbf{3.96$\pm$0.05}&\textbf{2.68 $\pm$0.09}&\textbf{3.63$\pm$0.10}&\textbf{19.14$\pm$0.20}&\textbf{35.26$\pm$0.10}\\
\hline
\end{tabular}
}
\caption{Error rate (\%) of the testing methods with PreAct ResNet-18~\cite{DBLP:journals/corr/HeZR016} as baseline.  We report mean scores over 5 runs with standard deviations (denoted $\pm$). 
Best results are  in \textbf{Bold}. 
  }    \label{tab:accuracy:resnet18} 
\end{table*} 

\begin{table*}[h]
  \centering
\scalebox{0.92}{
\begin{tabular}{l|c|c|c|c|c|c}\hline
Methods& MNIST&Fashion&SVHN&Cifar10&Cifar100&Tiny-ImageNet\\ \hline
ResNet-50 &0.61$\pm$0.05
	&4.55$\pm$0.14	&3.22$\pm$0.05&4.83$\pm$0.30&23.10$\pm$0.62	&35.67$\pm$0.50\\
ULS-0.1 &0.63$\pm$0.02&4.58$\pm$0.16&2.98$\pm$0.02&4.98$\pm$0.25&23.90$\pm$0.99&35.02$\pm$0.39\\
ULS-0.2 &0.62$\pm$0.03&4.52$\pm$0.04&3.08$\pm$0.03&5.00$\pm$0.35&23.88$\pm$0.73&36.19$\pm$0.66\\
ULS-0.3 &0.65$\pm$0.03&4.51$\pm$0.15&3.04$\pm$0.07&5.16$\pm$0.16&23.17$\pm$0.50&36.14$\pm$0.06\\
\hline
Mixup  &0.57$\pm$	0.03&4.31$\pm$0.05	&2.85$\pm$0.07&4.29$\pm$0.28&19.48$\pm$0.48&32.36$\pm$0.53\\
Mixup-ULS0.1  &0.60$\pm$0.04	&4.28$\pm$0.12	&2.90$\pm$0.09&4.02$\pm$0.27&21.58$\pm$0.86&32.11$\pm$0.09\\
Mixup-ULS0.2  &0.58$\pm$0.02	&4.33$\pm$0.09	&2.89$\pm$0.07&4.09$\pm$0.10&20.87$\pm$0.51&32.81$\pm$0.48\\
Mixup-ULS0.3  &0.57$\pm$0.04	&4.29$\pm$0.11	&2.84$\pm$0.10&4.19$\pm$0.18&21.64$\pm$0.41&33.94$\pm$0.56\\
\hline
PLS &\textbf{0.51$\pm$
0.02}&\textbf{4.15$\pm$0.09}&\textbf{2.36$\pm$0.03}&\textbf{3.60$\pm$0.18}&\textbf{18.65$\pm$1.08}&\textbf{30.73$\pm$0.20}\\
\hline
\end{tabular}
}
\caption{Error rate (\%) of the testing methods with  ResNet-50~\cite{DBLP:journals/corr/HeZR016} as baseline.  We report mean scores over 5 runs with standard deviations (denoted $\pm$). 
Best results are  in \textbf{Bold}. 
  }    \label{tab:accuracy:resnet50} 
\end{table*} 

\section{Experiments}
We first  show our  method's superior  accuracy, and then visualize its high uncertainty training labels and  conservative classifications. 
\subsection{Datasets, Baselines, and Settings}

We use 
six benchmark image classification tasks.  
\noindent \textbf{MNIST} is a digit (1-10)  recognition dataset with 60,000 training and 10,000 test 28x28-dimensional gray-level images. 
\noindent 
\textbf {Fashion} is an image recognition dataset with the same scale as
MNIST, containing 10 classes of fashion product pictures.
\noindent 
\textbf{SVHN} is the Google street view house numbers recognition data set. It has 73,257 digits, 32x32 color images for training, 26,032 for testing, and 531,131 additional, easier samples. Following literature, we did not use the additional images.
\noindent 
\textbf{Cifar10} is an image classification task with  10 classes, 
  50,000 training  and 10,000 test samples. 
\noindent 
\textbf{Cifar100} is similar to Cifar10 but with 100 classes and 600 images each.
\textbf{Tiny-ImageNet}~\cite{ChrabaszczLH17} 
has 
200 classes, each with 500 training 
and 50 test 64x64x3 images. 

We conduct experiments  using the popular benchmarking networks  PreAct ResNet-18 and ResNet-50~\cite{DBLP:journals/corr/HeZR016}. 
We  compare with the 
label smoothing methods~\cite{MullerKH19,Lukasik2020DoesLS} (denoted as ULS) with various smoothing coefficients (i.e., $\alpha$  in Equation~\ref{lsloss}), where ULS-0.1, ULS-0.2, and ULS-0.3  denote the smoothing coefficient of 0.1, 0.2, and 0.3, respectively. 
We also compare  with the input-pair based data augmentation method  Mixup~\cite{Mixup17}. 
We further compare with methods that stacking the ULS on top of Mixup, denoted  Mixup-ULS0.1, Mixup-ULS0.2, and Mixup-ULS0.3. 

For Mixup, we use the authors' code at~\footnote{https://github.com/facebookresearch/mixup-cifar10} and the uniformly selected mixing coefficients between [0,1]. 
For PreAct ResNet-18 and ResNet-50, we use the  implementation  from Facebook~\footnote{https://github.com/facebookresearch/mixup-cifar10/models/}. For PLS, the added fully connected layer is the same as  the last fully connected layer of the baseline network with a Sigmoid function on the top. 
All models  are trained using mini-batched (128 examples) backprop, 
 with the  {\em exact settings} as in the Facebook codes,     for 400 epochs.   Each reported  value (accuracy or error rate) is the mean of   five runs  on a NVIDIA GTX TitanX GPU with 12GB memory. 

\subsection{Predictive Accuracy} 
The  error rates obtained by  ULS, Mixup, Mixup-ULS, and PLS using    ResNet-18  as baseline on the six test datasets are presented in Table~\ref{tab:accuracy:resnet18}. The  results with ResNet-50 as baselines  are provided in Table~\ref{tab:accuracy:resnet50}. 

Table~\ref{tab:accuracy:resnet18} shows that PLS outperforms, in terms of predictive error, the  ResNet-18, the label smoothing models (ULS-0.1, ULS-0.2, ULS-0.3),  Mixup, stacking ULS on top of Mixup (Mixup-ULS0.1, Mixup-ULS0.2, Mixup-ULS0.3) on  all the six datasets. 
For example, the relative improvement of PLS over ResNet-18   on  Cifar10 and MNIST  
are over 30\% and 24\%, respectively. 
When considering PLS and the average error obtained by the three ULS models, 
on both  Cifar10 and MNIST, the relative improvement is over 23\%. 
It is also promising to see that on  Tiny-ImageNet,  PLS  reduced the absolute error rates over  Mixup, Mixup-ULS, and ULS from about 38\% to 35\%. 

For the cases with ResNet-50 as baselines, 
Table~\ref{tab:accuracy:resnet50}  indicates that similar error reductions are obtained by PLS. Again, on all the  testing datasets, PLS outperforms all the comparison baselines. For example, for  Cifar10, the relative improvement achieved by PLS over   ResNet-50 and the average of the three label smoothing strategies (i.e. ULS) are 25.47\% and 28.67\%, respectively. 
For  Cifar100 and Tiny-ImageNet, PLS reduced the absolute error rates of ULS from  23\% and 36\% to 18.65\% and 30.73\% respectively.

An important observation here is that, 
 stacking label smoothing on top of Mixup (i.e., Mixup-ULS) did not improve, or even degraded, Mixup's  accuracy. For example, for Cifar10, Cifar100, and Tiny-ImageNet (the last three columns in  Tables~\ref{tab:accuracy:resnet18} and~\ref{tab:accuracy:resnet50}),     Mixup-ULS obtained similar or slightly higher error rate than  Mixup. The reason here is that  Mixup creates samples with soft labels through linearly interpolating between [0, 1], which is a form of label smoothing regularization~\cite{carratino2020mixup}. Consequently,  stacking  another label smoothing regularizer on   top of Mixup can easily mess up the soft training targets,  resulting in underfitting. 
  Promisingly,  PLS  was able to improve over Mixup and Mixup-ULS. For example, on  Tiny-ImageNet PLS outperformed Mixup and Mixup-ULS by reducing the error  from 38.06\% to 35.26\% and from 32.36\% to 30.73\%, respectively, when using  ResNet-18 and ResNet-50. Similar error reduction can be observed on  Cifar10 and Cifar100. 

\subsection{Ablation Studies}
\textbf{Impact of Learned Distribution and Original  Samples}   
\label{ablation}
We   evaluate the impact of the key components in  PLS using  ResNet-18 and ResNet-50 on Cifar100. Results are in Table~\ref{aaaablation}. The key components include  1) removing the  learned smoothing distribution mass in Equation~\ref{merge},  2) excluding the original training inputs as discussed in the method section, and 3) replacing the learned smoothing distribution mass with uniform distribution with weight coefficients of 0.1, 0.2, and 0.3 (denoted  UD-0.1, UD-0.2 and UD-0.3). 

The error rates obtained in  Table~\ref{aaaablation} show that, 
  both  the learned smoothing distribution  and  the original training samples are critical for  PLS. 
In particular, when excluding the original  samples  from  training, the predictive error of PLS  dramatically increased from about 19\% to nearly 24\% for both  ResNet-18 and ResNet-50. The reason,  as discussed in the method section, 
 is that,  
without the original training samples, the networks are trained with midpoint samples only, thus may lack  information on the  validation samples with one-hot labels.

\begin{table}[h]
  \centering
  \begin{tabular}{l|r|r}\hline
 \multirow{2}{*}{PLS }& ResNet-18&ResNet-50 \\ 
  &19.14&18.65 \\ \hline
----  no learned distribution  &21.06&19.35\\
----  no original  samples  &23.84&24.42\\ 
 \hline
----  UD 0.1&19.50&18.91\\
----  UD 0.2&19.25&18.81\\
----  UD 0.3&19.31&18.89
\\ 
\hline
\end{tabular}
  \caption{Error rates (\%) on Cifar100 by PLS  while varying its key components: no learned  distribution, no original  samples, replacing learned smoothing distribution with uniform  distribution.  
\label{aaaablation}}

\end{table}

Also,  Table~\ref{aaaablation} indicates that,  replacing the learned smoothing distribution  in PLS with manually tuned Uniform distribution  (i.e., UD) obtained slightly larger errors. This  indicates that  PLS  is able to  learn the  distribution  to smooth the two target labels in the midpoint samples, resulting in superior  accuracy and excluding the need for the  coefficient search  for different  applications.

\textbf{Learning Smoothing Coefficient}
We also conducted experiments of  learning the smoothing coefficient, instead of learning the smoothing distribution, for each midpoint example. That is,  we replace Equation~\ref{sha}  with $\sigma(W_{t} S_{ij})$ where $W_{t}$ is $R^{1 \times m}$ instead of $R^{K \times m}$. Table~\ref{coefficientpredictin} presents the error rates (\%) obtained using PreAct ResNet-18 and ResNet-50 on Cifar100 and Cifar10. 
These results suggest that learning to predict smoothing distribution significantly better than predicting the  smoothing coefficient in PLS.

\begin{table}[h]
  \centering
  \begin{tabular}{l|r|r}\hline
 \multirow{2}{*}{PLS }& ResNet-18&ResNet-50 \\ 
  & pred. coeff. / PLS& pred. coeff. / PLS \\ \hline
Cifar100  &21.60 / 19.14&19.21 / 18.65\\
Cifar10 &4.67 / 3.63&4.06 / 3.60\\ 
\hline
\end{tabular}
  \caption{Error rates (\%) on Cifar100 and Cifar10 obtained by PLS  while predicting the smoothing coefficient (pred. coeff.) instead of learning the smoothing distribution (PLS) for each sample pair.  
\label{coefficientpredictin}}

\end{table}

\begin{figure}[h]
\centering
    \includegraphics[width=3.20972in, height=1.3in
    ]{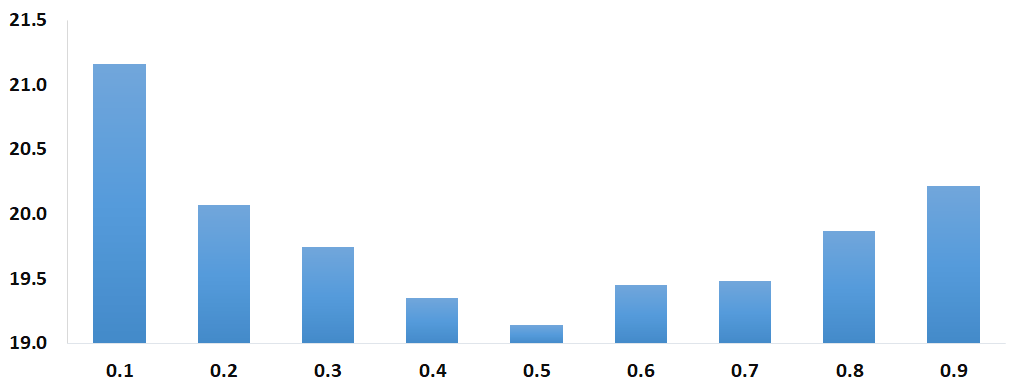}     
  \caption{
  Error rates (\%, Y-axis) on Cifar100 obtained by 
  varying the weight factor in PLS from 0.1 to 0.9 (X-axis). } 
  \label{weighting}
\end{figure}

\textbf{Re-weight Smoothing Strength}
PLS distributes half of a midpoint sample's  ground truth distribution mass over the non-ground truth targets (Equation~\ref{merge}). 
We here  evaluate the impact of different weight factors between the ground truth and non-ground truth targets, by varying it from 0.1 to 0.9 (0.5 equals to the average used by PLS). The results obtained by PLS  using  ResNet-18 on Cifar100 are in Figure~\ref{weighting}. The error rates  in Figure~\ref{weighting} suggest 
that average used by  midpoints  in  PLS 
provides better accuracy than other weighting ratios. This is expected as discussed in Section~\ref{evovinglabel}. 

\subsection{ Uncertainty  Label and Conservative Classification}
\label{confidence}
This section will show  that  PLS utilizes  high uncertainty  labels from midpoint samples for training and produces very conservative classification decisions in testing.
\subsubsection{High Uncertainty  Labels in Training} 
\begin{figure}[h]
\caption{
Average distribution  of ground truth targets 
(green bars)
and the top 5 largest  non-ground truth targets (orange bars) used by PLS in  training  with  ResNet-18 on Cifar100 (top) and Cifar10 (bottom).  
X-axis is the classes and Y-axis the  distribution mass. 
 }
\label{fig:top3}
\centering
\includegraphics[
width=3.50972in
]{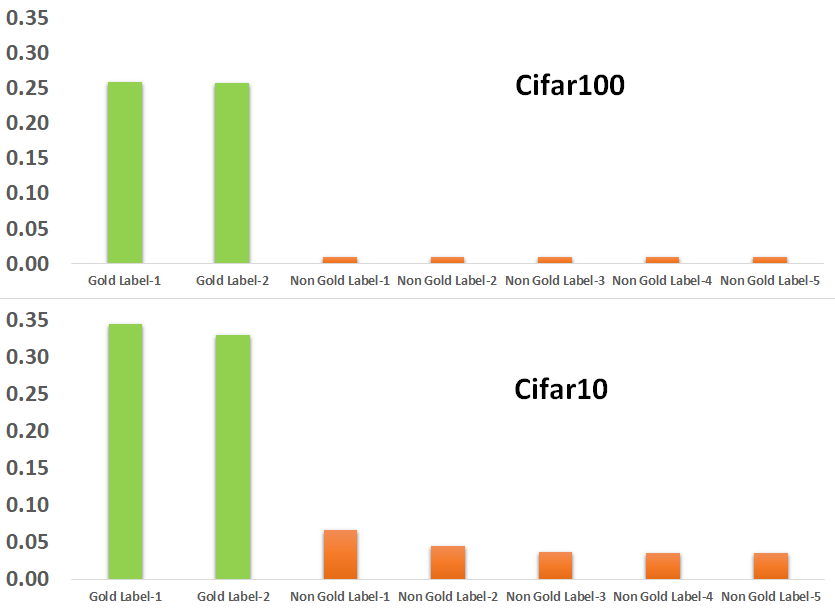}

\end{figure}

 We visualize, in Figure~\ref{fig:top3}, the soft target labels used for training by  PLS  with ResNet-18 on  Cifar100  (top) and Cifar10 (bottom). Figure~\ref{fig:top3} depicts the   soft label values of the training samples for both the ground truth targets (in green) and their top 5 largest  non-ground truth classes (in orange). The figure presents the average values over  all the training samples resulting from sample pairs with two different one-hot true labels.  Here,   X-axis depicts the training targets, and   Y-axis  the corresponding distribution mass. 
 
Results in Figure~\ref{fig:top3} indicate that, PLS uses much smaller target values for the ground truth labels during training, when compared to the one-hot representation label used by the baselines and the soft targets used by label smoothing ULS and Mixup. 
For example, the largest ground truth  targets for PLS are around  0.25 and 0.35  (green bars), respectively, for Cifar100 and Cifar10. These values are much smaller than 1.0 used by the baselines and  0.9 and 0.5 used by ULS-0.1 and Mixup respectively. 
Consequently, in PLS, the distance between  the  ground truth targets and the non-ground truth targets are much smaller than that in the baseline, ULS-0.1 and Mixup. In addition, the distance between the two ground truth targets in PLS (green bars) is very small, when compared to the distance between the ground truth and non-ground truth target values (green vs. orange bars).

These results indicate that 
the training samples in PLS have smoother training targets across  classes, and  those  training signals are far from 1.0 (thus with much higher uncertainty), which in turn impacts how PLS makes its classification decisions as will be discussed next.  

\subsubsection{Low Predicted Softmax Scores in Testing} 
\begin{figure}[h]
\caption{ Histograms of predicted softmax scores on the validation data   by  ResNet-18 (top row), ULS-0.1 (second row), Mixup (third row), and PLS (bottom row) on Cifar100 (left) and Cifar10 (right). 
 }
	\centering
\includegraphics[width=4.3814in
]{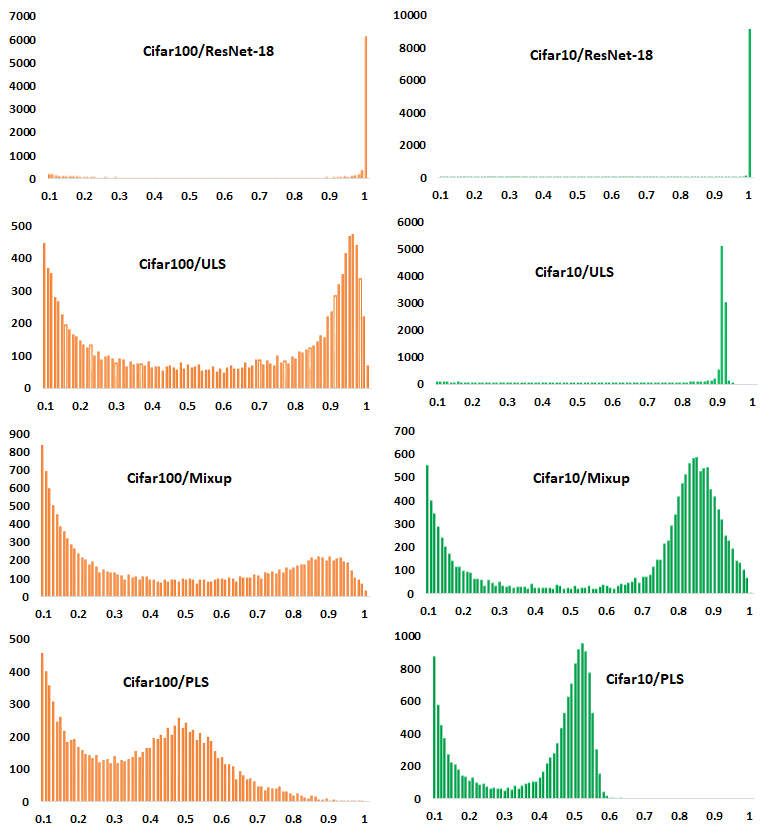}
	\label{fig:hist}
\end{figure}

One effect of the high uncertainty  training signals (far from 1.0) as discussed above was reflected on the model's prediction scores made in test time. 
Figure~\ref{fig:hist}  visualizes the predicted winning softmax scores  made by  ResNet-18 (top row), ULS-0.1 (second row), Mixup (third row) and PLS (bottom row) on all the 10K test data samples in Cifar100 (left column) and Cifar10 (right column). To have better visualization, we have removed the predictive scores less than 0.1 for all the  methods since all models obtained similar results for confidences smaller than 0.1. 

For  Cifar100, results on the left of  Figure~\ref{fig:hist} indicate that    ResNet-18 produced very confident predictions (top-left subfigure), namely skewing large mass of its predicted softmax scores on 1.0 (i.e., 100\% confidence). On the other hand,   ULS and Mixup  were able to 
decrease its prediction confidence at test time (middle rows/left). They  spread their predictive confidences to the two ends, namely moving most of the predicted  scores into two bins [0.1-0.2] and [0.8-1.0]. 
In contrast,  PLS  produced very conservative predicted softmax scores, by distributing many of its predicted scores to the middle, namely 0.5, and with sparse distribution for scores  larger than 0.7 (bottom left subfigure).

For Cifar10, results  on the right of Figure~\ref{fig:hist} again indicate that  ResNet-18 (top) produced very confident predictions,  putting large mass of its predicted softmax scores  near 1.0. 
For  ULS and Mixup  (middle rows), the predicted softmax scores were also distributed near the 1.0, but they are much less than that of  ResNet-18. In contrast,  PLS  (bottom right) again generated very conservative predicted softmax scores. Most of them distributes near the middle  of the  softmax score range, namely 0.5, with a very few larger than 0.6. 

These results suggest that, resulting from high uncertainty  training  signals across classes, PLS becomes extremely conservative when generating predicted scores in test time, producing low  winning softmax scores for  classification. 

\subsubsection{Impact on  Model Calibration}
\label{tempscaling}
The  conservative predictions generated by PLS improve the model's accuracy, but  how such low softmax scores  affect the calibration of the output probabilities? 
In this section, we report the Expected Calibration Error (ECE)~\cite{10.5555/3305381.3305518} obtained by the baseline PreAct ResNet-18, ULS-0.1, Mixup and PLS on the test set with 15 bins as used in~\cite{MullerKH19} for both  Cifar100 and Cifar10. 

\begin{figure}[h]
\caption{ ECE curves on the   validation data of Cifar100 (left) and Cifar10 (right)  from  ResNet-18, ULS-0.1, Mixup, PLS and PLS with TS. X-axis is the training epoch and Y-axis  the ECE value.}
	\centering
\includegraphics[width=4.3in]{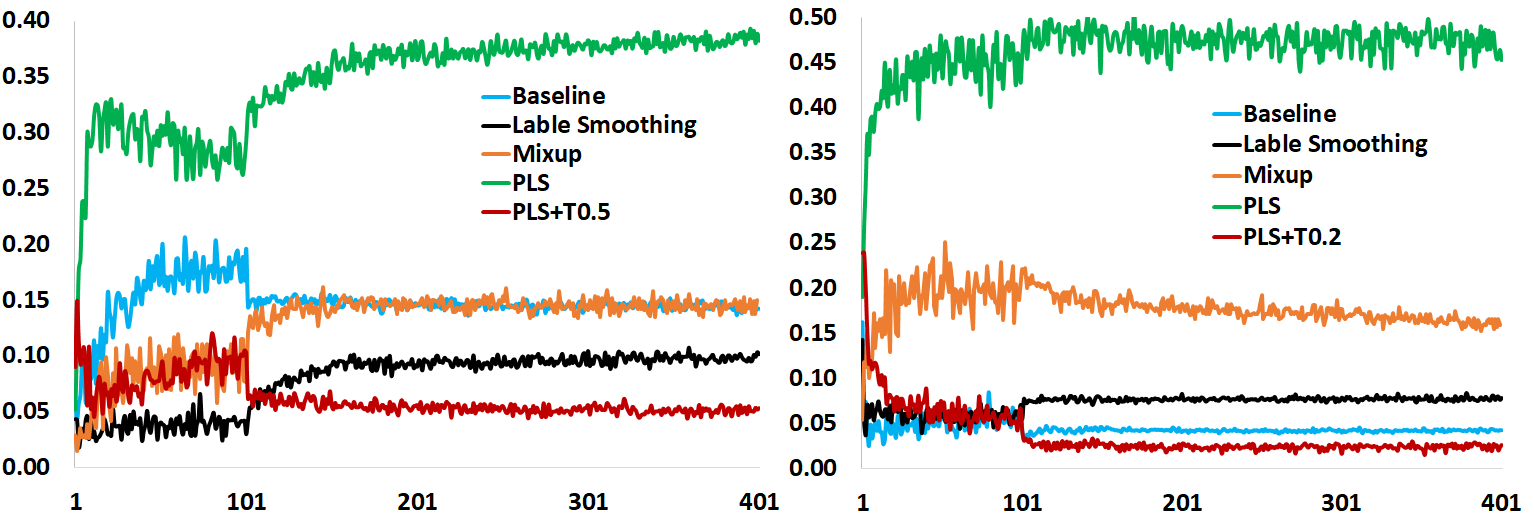}
	\label{fig:ece}
\end{figure}

Results in Figure~\ref{fig:ece} indicate that ULS (dark curve) is able to reduce the miscalibration error ECE on the Cifar100 data set (left subfigure), but for the Cifar10 dataset (right subfigure), ULS  has larger ECE error after 100 epochs of training than the baseline model. Also, Mixup has higher ECE error than ULS on both Cifar100 and Cifar10. However, the ECE errors obtained by the PLS methods for both the Cifar100 and Cifar10 are much larger than   the baseline,  ULS and Mixup. 
Note that, although the authors in~\cite{10.5555/3305381.3305518} state that the Batch Normalization (BN) strategy~\cite{DBLP:conf/nips/Ioffe17} also increases the miscalibration ECE errors for unknown reasons, we doubt that  the PLS will have the same reason as the BN approach. 
 This is because the main characteristic of the PLS model is that it produces extremely conservative winning softmax scores which is not the case for the BN strategy.  
We here suspect that the high ECE score of the PLS method may be caused by the fact that ECE is an evenly spaced binning metrics but the PLS produces sparse dispersion of the softmax scores  across the range. 

To verify the above hypothesis, we further investigate the Temperature Scaling (TS) method~\cite{10.5555/3305381.3305518}, which enables redistributing the distribution dispersion after training with no impact on the testing accuracy. 
During testing, TS  multiplies the logits by a scalar before applying the softmax operator. 
 We apply this TS technique to PLS, and present the results in Figure~\ref{fig:ece}, depicting by  red curve in the left and right subfigures for the Cifar100 and Cifar10, respectively. The TS factors was 0.5 and 0.2 respectively for Cifar100 and Cifar10, which were found by a search with 10 percentage of the training data. 
Figure~\ref{fig:ece} indicates that  TS can significantly improve the  calibration of  PLS for both cases. The ECE errors obtained by PLS-TS for both  Cifar100 and Cifar10 (red curves) are lower than that of the baseline,  ULS, and Mixup.

\subsection{Testing on Out-of-Distribution Data}
Section~\ref{confidence} shows that PLS produces very conservative classification decisions for  in-distribution samples. We here
 explore the effect of PLS on out-of-distribution data.  

\begin{figure}[h]
\caption{ 
Winning softmax scores produced by PLS, Mixup, ULS, and baseline models, when being trained with Cifar10 (left) and Cifar100 (right) and then testing on SVHN with PreAct ResNet-18 (top) and ResNet-50 (bottom). 
}%
	\centering
\includegraphics[width=4.3in]{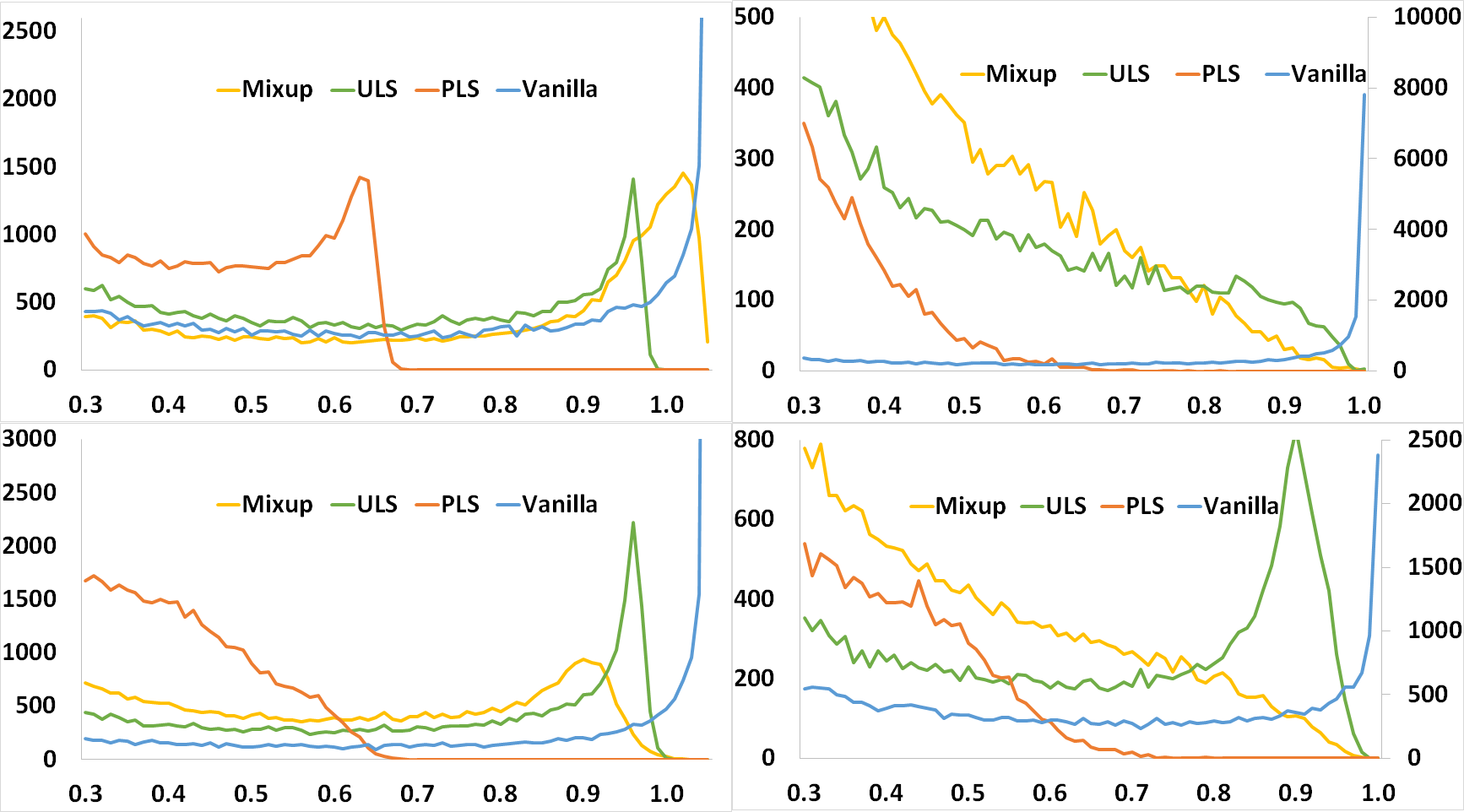}
	\label{fig:ood}
\end{figure}

In this experiment, we first train a  ResNet-18 or ResNet-50 network (denoted as vanilla models) with   in-distribution data (using either Cifar10 or Cifar100) and
then let the trained networks to predict on the testing samples from the SVHN dataset (i.e., out-of-distribution samples). 
We compare our method with the vanilla baseline (ResNet-18 or ResNet-50), Mixup, and ULS-0.1. The winning predicted softmax scores on the SVHN testing samples  are presented in  Figure~\ref{fig:ood}, where the top and bottom rows depict the results for ResNet-18 and ResNet-50, respectively. 

Figure~\ref{fig:ood} shows that, 
PLS again produces low winning predicted softmax scores (namely less confident) against the samples from the SVHN dataset when  training with either Cifar10 or Cifar100 data, when compared to  Mixup, ULS, and the vanilla models. 
For example, when being trained with Cifar10 and tested on the SVHN data with ResNet-18 (the top-left subfigure), PLS  produced  (orange curve) a spike of score distribution  near the midpoint 0.5, with extremely spare distribution in the regions of high confidence.  
In contrast, while Mixup and ULS (yellow/green curves) are more conservative than
the vanilla model on out-of-distribution data, it is noticeably  more overconfident than the PLS strategy by producing a  spike of prediction with 90\% confidence at the right end of the figure.

\section{Conclusion and Future Work}
We proposed a novel output distribution  regularization technique, coined PLS, which learns smoothing distribution for midpoint samples that average random sample pairs. 
 We empirically showed   PLS'  superior   accuracy over label smoothing and Mixup models. 
 We  visualized   the high uncertainty training labels of  the  midpoint samples, which cause PLS to produce  very  low winning softmax scores  for unseen  in and out of distribution samples. 


Our studies here suggest some  interesting directions for future investigation. For example, what are the 
other  benefits arising from  high uncertainty training labels and  conservative classification?  
Would such uncertain predictions help with  beam search or ranking process for some downstream applications? 
Does the proposed method benefit from mixing  more than two samples for smoothing?  
Another interesting research direction would be providing theoretical explanation  on 
 why the method works considering that the synthetic images are not realistic. 
We are also interested in applying our strategy to other domains  beyond image. 

\bibliographystyle{splncs04}
\bibliography{ECML_PKDD_2021}

\begin{thebibliography}{10}
\providecommand{\url}[1]{\texttt{#1}}
\providecommand{\urlprefix}{URL }
\providecommand{\doi}[1]{https://doi.org/#1}

\bibitem{ahn2019variational}
Ahn, S., Hu, S.X., Damianou, A., Lawrence, N.D., Dai, Z.: Variational
  information distillation for knowledge transfer. In: arXiv (2019)

\bibitem{DBLP:journals/corr/abs-1906-06875}
Archambault, G.P., Mao, Y., Guo, H., Zhang, R.: Mixup as directional
  adversarial training. vol. abs/1906.06875 (2019)

\bibitem{carratino2020mixup}
Carratino, L., Cissé, M., Jenatton, R., Vert, J.P.: On mixup regularization.
  In: arXiv (2020)

\bibitem{Chorowski2016TowardsBD}
Chorowski, J., Jaitly, N.: Towards better decoding and language model
  integration in sequence to sequence models. In: INTERSPEECH (2016)

\bibitem{ChrabaszczLH17}
Chrabaszcz, P., Loshchilov, I., Hutter, F.: A downsampled variant of imagenet
  as an alternative to the {CIFAR} datasets. In: arXiv (2017)

\bibitem{Furlanello2018BornAN}
Furlanello, T., Lipton, Z.C., Tschannen, M., Itti, L., Anandkumar, A.: Born
  again neural networks. In: ICML (2018)

\bibitem{10.5555/3305381.3305518}
Guo, C., Pleiss, G., Sun, Y., Weinberger, K.Q.: On calibration of modern neural
  networks. In: ICML. ICML’17, JMLR.org (2017)

\bibitem{DBLP:conf/aaai/Guo20}
Guo, H.: Nonlinear mixup: Out-of-manifold data augmentation for text
  classification. In: AAAI. pp. 4044--4051 (2020)

\bibitem{DBLP:journals/corr/abs-1905-08941}
Guo, H., Mao, Y., Zhang, R.: Augmenting data with mixup for sentence
  classification: An empirical study. vol. abs/1905.08941 (2019)

\bibitem{adamixup}
Guo, H., Mao, Y., Zhang, R.: Mixup as locally linear out-of-manifold
  regularization. In: AAAI. pp. 3714--3722 (2019)

\bibitem{DBLP:journals/corr/HeZR016}
He, K., Zhang, X., Ren, S., Sun, J.: Identity mappings in deep residual
  networks. ECCV  (2016)

\bibitem{hinton2015distilling}
Hinton, G., Vinyals, O., Dean, J.: Distilling the knowledge in a neural
  network. In: arXiv (2015)

\bibitem{Huang2019GPipeET}
Huang, Y., Cheng, Y., Chen, D., Lee, H., Ngiam, J., Le, Q.V., Chen, Z.: Gpipe:
  Efficient training of giant neural networks using pipeline parallelism. In:
  NeurIPS (2019)

\bibitem{DBLP:conf/nips/Ioffe17}
Ioffe, S.: Batch renormalization: Towards reducing minibatch dependence in
  batch-normalized models. In: Guyon, I., von Luxburg, U., Bengio, S., Wallach,
  H.M., Fergus, R., Vishwanathan, S.V.N., Garnett, R. (eds.) NeurIPS. pp.
  1945--1953 (2017)

\bibitem{structurels}
Li, W., Dasarathy, G., Berisha, V.: Regularization via structural label
  smoothing. In: AISTAT (2020)

\bibitem{Lukasik2020DoesLS}
Lukasik, M., Bhojanapalli, S., Menon, A.K., Kumar, S.: Does label smoothing
  mitigate label noise? ICML  (2020)

\bibitem{mobahi2020selfdistillation}
Mobahi, H., Farajtabar, M., Bartlett, P.L.: Self-distillation amplifies
  regularization in hilbert space. In: arXiv (2020)

\bibitem{MullerKH19}
M{\"{u}}ller, R., Kornblith, S., Hinton, G.E.: When does label smoothing help?
  In: NIPS (2019)

\bibitem{DBLP:conf/iclr/PereyraTCKH17}
Pereyra, G., Tucker, G., Chorowski, J., Kaiser, L., Hinton, G.E.: Regularizing
  neural networks by penalizing confident output distributions. In: ICLR
  workshop (2017)

\bibitem{Real2019RegularizedEF}
Real, E., Aggarwal, A., Huang, Y., Le, Q.V.: Regularized evolution for image
  classifier architecture search. In: AAAI (2019)

\bibitem{Szegedy2016RethinkingTI}
Szegedy, C., Vanhoucke, V., Ioffe, S., Shlens, J., Wojna, Z.: Rethinking the
  inception architecture for computer vision. 2016 IEEE Conference on Computer
  Vision and Pattern Recognition (CVPR) pp. 2818--2826 (2016)

\bibitem{DBLP:conf/iclr/TokozumeUH18}
Tokozume, Y., Ushiku, Y., Harada, T.: Learning from between-class examples for
  deep sound recognition. In: ICLR (2018)

\bibitem{Vaswani2017AttentionIA}
Vaswani, A., Shazeer, N., Parmar, N., Uszkoreit, J., Jones, L., Gomez, A.N.,
  Kaiser, L., Polosukhin, I.: Attention is all you need. ArXiv
  \textbf{abs/1706.03762} (2017)

\bibitem{7780883}
{Xie}, L., {Wang}, J., {Wei}, Z., {Wang}, M., {Tian}, Q.: Disturblabel:
  Regularizing cnn on the loss layer. In: 2016 IEEE Conference on Computer
  Vision and Pattern Recognition (CVPR). pp. 4753--4762 (2016)

\bibitem{DBLP:conf/aaai/YangXQY19}
Yang, C., Xie, L., Qiao, S., Yuille, A.L.: Training deep neural networks in
  generations: {A} more tolerant teacher educates better students. In: AAAI.
  pp. 5628--5635. {AAAI} Press (2019)

\bibitem{yuan2019revisit}
Yuan, L., Tay, F.E.H., Li, G., Wang, T., Feng, J.: Revisit knowledge
  distillation: a teacher-free framework. In: arXiv (2019)

\bibitem{cutmix19}
Yun, S., Han, D., Chun, S., Oh, S.J., Yoo, Y., Choe, J.: Cutmix: Regularization
  strategy to train strong classifiers with localizable features. In: ICCV. pp.
  6022--6031. {IEEE} (2019)

\bibitem{Mixup17}
Zhang, H., Ciss{\'{e}}, M., Dauphin, Y.N., Lopez{-}Paz, D.: mixup: Beyond
  empirical risk minimization. In: ICLR (2018)

\bibitem{DBLP:conf/aaai/ZhuJZGHSZ20}
Zhu, Z., Jiang, X., Zheng, F., Guo, X., Huang, F., Sun, X., Zheng, W.:
  Viewpoint-aware loss with angular regularization for person
  re-identification. In: AAAI. pp. 13114--13121 (2020)

\bibitem{8579005}
{Zoph}, B., {Vasudevan}, V., {Shlens}, J., {Le}, Q.V.: Learning transferable
  architectures for scalable image recognition. In: 2018 IEEE/CVF Conference on
  Computer Vision and Pattern Recognition. pp. 8697--8710 (2018)

\end{thebibliography}

\end{document}